\documentclass[11pt,a4paper]{article}
\usepackage{times,latexsym}
\usepackage{url}
\usepackage[T1]{fontenc}

\usepackage[acceptedWithA]{tacl2021v1}

\usepackage{rotating}
\usepackage[]{xcolor}
\usepackage{colortbl}
\usepackage{hhline}
\usepackage{color,array}
\definecolor{heat1}{RGB}{101,133,70} % 150-200
\definecolor{heat3}{RGB}{125,160,98} % 75-100
\definecolor{heat4}{RGB}{130,170,100} % 50-75
\definecolor{heat5}{RGB}{154,199,124} % 40-49
\definecolor{heat6}{RGB}{180,210,147} % 30-39 
\definecolor{heat7}{RGB}{190,220,170} % 20-29
\definecolor{heat8}{RGB}{215,235,190} % 11-19
\definecolor{heat10}{RGB}{235,245,225} % 6-10
\usepackage{arydshln}

\usepackage{xspace,mfirstuc,tabulary}

\newcommand{\ex}[1]{{\sf #1}}

\newcommand{\sid}[1]{{\begin{sideways} #1 \end{sideways}}}

\newif\iftaclinstructions
\taclinstructionsfalse % AUTHORS: do NOT set this to true
\iftaclinstructions
\renewcommand{\confidential}{}
\renewcommand{\anonsubtext}{(No author info supplied here, for consistency with
TACL-submission anonymization requirements)}
\newcommand{\instr}
\fi

\iftaclpubformat % this "if" is set by the choice of options

\else

\fi

\usepackage{graphicx}
\usepackage{linguex}
\usepackage{enumitem}
\usepackage{float}

\title{Design Choices for Crowdsourcing Implicit Discourse Relations: \\ Revealing the Biases Introduced by Task Design}

 \author{Valentina Pyatkin\textsuperscript{1} \,
 Frances Yung\textsuperscript{2} \,
 Merel C.J. Scholman\textsuperscript{2,3} \, \\
 {\bf Reut Tsarfaty\textsuperscript{1}\,
 \bf Ido Dagan\textsuperscript{1}\,
 Vera Demberg\textsuperscript{2}}\\
\textsuperscript{1}Bar Ilan University, Ramat Gan, Israel \\
\textsuperscript{2}Saarland University, Saarbr\"ucken, Germany \\
\textsuperscript{3}Utrecht University, Utrecht Netherlands\\
  {\tt  \{pyatkiv,reut.tsarfaty\}@biu.ac.il;
         dagan@cs.biu.ac.il}\\
  \tt  \{m.c.j.scholman,frances,vera\}@coli.uni-saarland.de}

\date{}

\begin{document}
\maketitle
\begin{abstract}

Disagreement in natural language annotation has mostly been studied from a perspective of biases introduced by the annotators and the annotation frameworks. Here, we
propose to analyze another source of bias: task design bias,
which has a particularly strong impact on crowdsourced linguistic annotations where natural language is used to elicit the interpretation of laymen annotators. 
For this purpose we look at implicit discourse relation annotation, a task that has repeatedly been shown to be difficult due to the relations' ambiguity. We compare the annotations of 1,200 discourse relations obtained using two distinct annotation tasks and quantify the biases of both methods across four different domains. Both methods are natural language annotation tasks designed for crowdsourcing.
We show that the task design can push annotators towards certain relations and that some discourse relations senses can be better elicited with one or the other annotation approach.
We also conclude that this type of bias should be taken into account when training and testing models.
\end{abstract}

\section{Introduction}
\label{sec:introduction}
Crowdsourcing has become a popular method for data collection. It not only allows researchers to collect large amounts of annotated data in a shorter amount of time, but also captures human inference in natural language, which should be the goal of benchmark NLP tasks \cite{manning20061}.
In order to obtain reliable annotations, the crowdsourced labels are traditionally aggregated to a single label per item, using simple majority voting or annotation models that reduce noise from the data based on the disagreement among the annotators \cite{hovy2013,passonneau2014}.
However, there is increasing consensus that disagreement in annotation cannot be generally discarded as noise in a range of NLP tasks, such as natural language inferences \cite{de2012did,pavlick2019inherent,chen2019uncertain,nie2020can}, word sense disambiguation \cite{jurgens2013embracing}, question answering \cite{min2020ambigqa,ferracane2021did}, anaphora resolution \cite{poesio2005reliability,poesio2006underspecification}, sentiment analysis \cite{diaz2018addressing,cowen2019mapping} and stance classification \cite{waseem2016you,luo2020detecting}.
Label distributions are proposed to replace categorical labels in order to represent the label ambiguity \cite{aroyo2013,pavlick2019inherent,uma2021learning,dumitrache2021empirical}.

\begin{figure}[t]
    \centering
    \includegraphics[width=\linewidth]{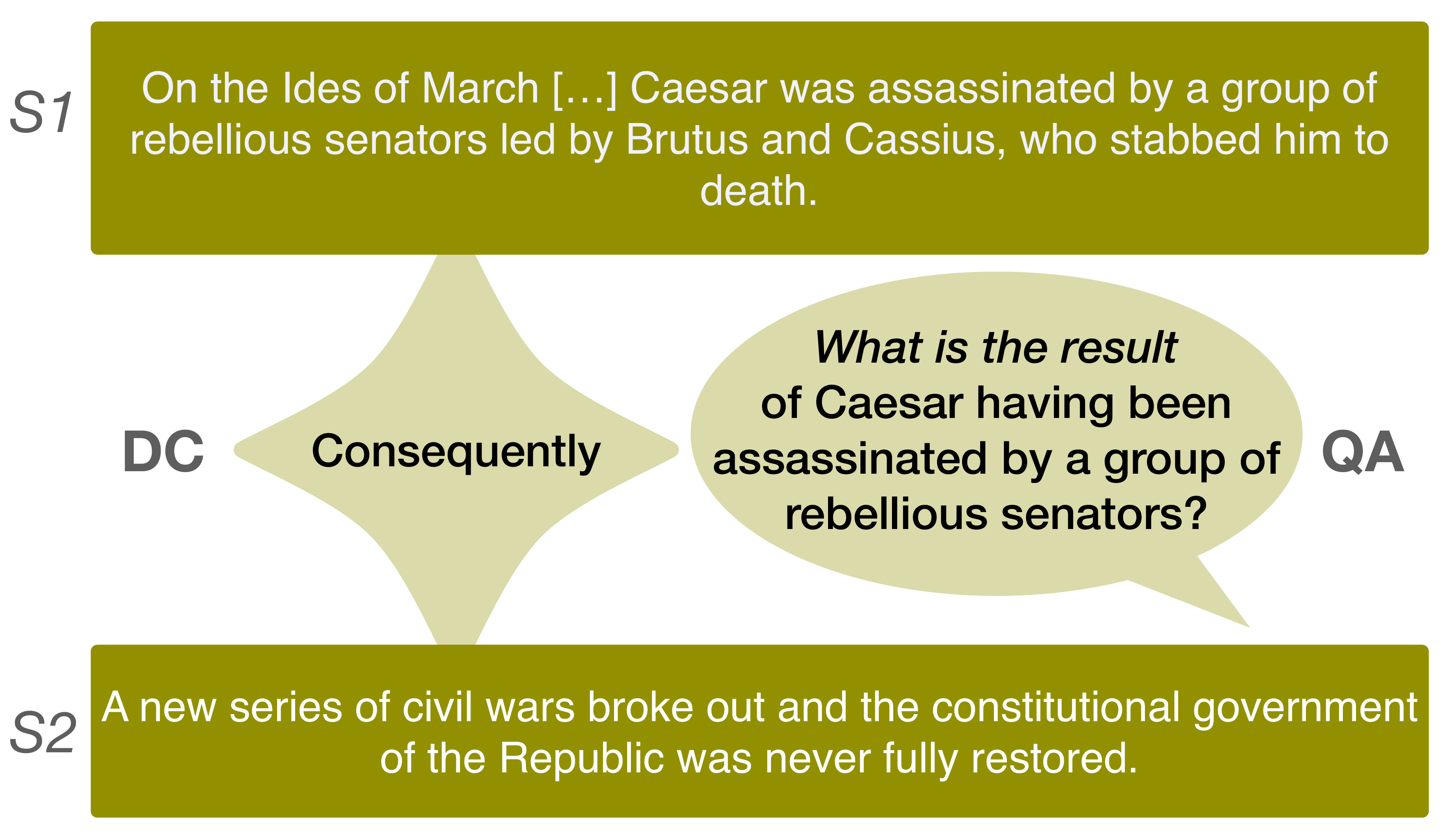}
    \vspace{-1em}
    \caption{Example of two relational arguments (S1 and S2) and the DC and QA annotation in the middle.}
    \label{fig:fig1}
\end{figure}

There are various reasons behind the ambiguity of linguistic annotations 
\cite{dumitrache2015crowdsourcing,jiang2022investigating}. 
\citet{aroyo2013} summarize the sources of ambiguity into three categories: the text, the annotators, and the annotation scheme. In downstream NLP tasks, it would be helpful if models could detect possible alternative interpretations of ambiguous texts, or predict a distribution of interpretations by a population.
In addition to the existing works on the disagreement due to annotators' bias, the effect of annotation frameworks has also been studied, such as the discussion on whether entailment should include pragmatic inferences \cite{pavlick2019inherent}, the effect of the granularity of the collected labels \cite{chung2019efficient}, or the system of labels that categorize the linguistic phenomenon \cite{demberg2019compatible}.
In this work, we examine the effect of task design bias, which is independent of the annotation framework, on the quality of crowdsourced annotations.  
Specifically, we look at inter-sentential implicit discourse relation (DR) annotation, i.e., semantic or pragmatic relations between two adjacent sentences without a discourse connective to which the sense of the relation can be attributed.
Fig.~\ref{fig:fig1} shows an example of an implicit relation that can be annotated as \textit{Conjunction} or \textit{Result}.

Implicit DR annotation is arguably the hardest task in discourse parsing.
Discourse coherence is a feature of the mental representation that readers form of a text, rather than of the linguistic material itself \citep{sanders1992}. Discourse annotation thus relies on annotators’ interpretation of a text. Further, relations can often be interpreted in various ways \citep{rohde2016}, with multiple valid readings holding at the same time. These factors make discourse relation annotation, especially for implicit relations,  
a particularly difficult task. 
We collect 10 different annotations per DR, thereby focusing on distributional representations, which are more informative than categorical labels. 

Since DR annotation labels are often abstract terms that are not easily understood by laymen, we focus on ``natural language" task designs. Decomposing and simplifying an annotation task, where the DR labels can be obtained indirectly from the natural language annotations, has been shown to work well for crowdsourcing \citep{chang2016linguistic,scholman2017context,pyatkin2020}. 
Crowdsourcing with natural language has become increasingly popular. This includes tasks such as NLI \cite{bowman2015large}, SRL \cite{fitzgerald2018large} and QA \cite{rajpurkar2018know}. This trend is further visible in modeling approaches which cast traditional structured prediction tasks into NL tasks, such as for co-reference \cite{aralikatte2019ellipsis}, discourse comprehension \cite{ko2021discourse} or bridging anaphora \cite{hou2020bridging, elazar2022text}. It is therefore of interest to the broader research community to see how task design biases can arise, even when the tasks are more accessible to laymen.

We examine two distinct natural language crowdsourcing discourse relation annotation tasks (Fig.~\ref{fig:fig1}): \citet{yung2019} derive relation labels from discourse connectives (DC) that crowd workers insert; \citet{pyatkin2020} derive labels from Question Answer (QA) pairs that crowd workers write. Both task designs employ natural language annotations instead of labels from a taxonomy.
The two task designs, DC and QA, are used to annotate 1,200 implicit discourse relations in 4 different domains. This allows us to explore how the task design impacts the obtained annotations, as well as the biases that are inherent to each method. To do so we showcase the difference of various inter-annotator agreement metrics on annotations with distributional and aggregated labels.

We find that both methods have strengths and weaknesses in identifying certain types of relations. We further see that these biases are also affected by the domain. In a series of discourse relation classification experiments, we demonstrate the benefits of collecting annotations with mixed methodologies, we show that training with a soft loss with distributions as targets improves model performance and we find that cross-task generalization is harder than cross-domain generalization.

The outline of the paper is as follows. We introduce the notion of task design bias and analyze its effect on crowdsourcing implicit DRs, using two different task designs (Sec.~\ref{sec:method}-\ref{sec:results}). Next, we quantify strengths and weaknesses of each method using the obtained annotations, and suggest ways to reduce task bias (Sec.~\ref{sec:method_bias_sources}). Then we look at genre specific task bias (Sec.~\ref{sec:genre}). Lastly, we demonstrate the task bias effect on DR classification performance (Sec.~\ref{sec:models}).

\section{Background}

\subsection{Annotation Biases}
Annotation tends to be an inherently ambiguous task, often with multiple possible interpretations and without a single ground truth \cite{aroyo2013}. An increasing amount of research has studied annotation disagreements and biases.

Prior studies have focused on how crowdworkers can be biased. Worker biases are subject to various factors, such as their educational or cultural background, or other demographic characteristics. 
\citet{prabhakaran2021releasing} point out that for more subjective annotation tasks, the socio-demographic background of annotators contributes to multiple annotation perspectives and argue that label aggregation obfuscates such perspectives. 
Instead, soft labels are proposed, such as the ones provided by the CrowdTruth method \cite{dumitrache2018}, which require multiple judgements to be collected per instance \cite{uma2021learning}. 
\citet{bowman2021will} suggests that annotations that are subject to bias from methodological artifacts should not be included in benchmark datasets.  In contrast, \citet{Basile2021WeNT} argues that all kinds of human disagreements should be predicted by NLU models and thus included in evaluation datasets.

In contrast to annotator bias, a limited amount of research is available on bias related to the formulation of the task.
\citet{jakobsen2022sensitivity} show that argument annotations exhibit widely different levels of social group disparity depending on which guidelines the annotators followed. 
Similarly, \citet{buechel2017emobank,buechel2017readers} study different design choices for crowdsourcing emotion annotations and show that the perspective that annotators are asked to take in the guidelines affects annotation quality and distribution.
\citet{jiang2017understanding} study the effect of workflow for paraphrase collection and found that examples based on previous contributions prompt workers to produce more diverging paraphrases. 
\citet{hube2019understanding} show that biased subjective judgment annotations can be mitigated by asking workers to think about responses other workers might give and by making workers aware of their possible biases. 
Hence, the available research suggests that task design can affect the annotation output in various ways.
Further research studied the collection of multiple labels:
\citet{jurgens2013embracing} compare between selection and scale rating and find that workers would choose an additional label for a word sense labelling task. In contrast, \citet{scholman2017context} find that workers usually opt not to provide an additional DR label even when allowed. \citet{chung2019efficient} compare various label collection methods including single / multiple labelling, ranking and probability assignment.
We focus on the biases in DR annotation approaches using the same set of labels, but translated into different "natural language" for crowdsourcing.

\subsection{DR annotation}
Various frameworks exist that can be used to annotate discourse relations, such as RST \cite{mann1988} and SDRT \cite{asher1993}.
In this work, we focus on the annotation of implicit discourse relations, following the framework used to annotate the Penn Discourse Treebank 3.0 \citep[PDTB, ][]{webber2019}. 
PDTB's sense classification is structured as a three-level hierarchy, with four coarse-grained sense groups in the first level and more fine-grained senses for each of the next levels.\footnote{We merge the belief and speech-act relation senses (which cannot be distinguished reliably by QA and DC) with their corresponding more general relation senses.}
The process is a combination of manual and automated annotation: an automated process identifies potential explicit connectives, and annotators then decide on whether the potential connective is indeed a true connective. If so, they specify one or more senses that hold between its arguments.  
If no connective or alternative lexicalization is present (i.e., for implicit relations), each annotator provides
one or more connectives that together express the sense(s) they infer.

DR datasets, such as PDTB \cite{webber2019}, RST-DT \cite{carlson2001} and TED-MDB \cite{zeyrek2019}, are commonly annotated by trained annotators, who are expected to be familiar with extensive guidelines written for a given task \cite{plank2014linguistically, artstein2008, riezler2014}. 
However, there have also been efforts to crowdsource discourse relation annotations \cite{kawahara2014,Kishimoto2018ImprovingCA,scholman2017context, pyatkin2020}. 
We investigate two crowdsourcing approaches that annotate inter-sentential implicit DRs and we deterministically map the NL-annotations to the PDTB3 label framework.

\subsubsection{Crowdsourcing DRs with the DC method}
\citet{yung2019} developed a crowdsourcing discourse relation annotation method using discourse connectives, referred to as the DC method. 
For every instance, participants first provide a connective which, in their view, best expresses the relation between the two arguments. Note that the connective chosen by the participant might be ambiguous. Therefore, participants disambiguate the relation in a second step, by selecting a connective from a list that is generated dynamically based on the connective provided in the first step.
When the first step insertion does not match any entry in the connective bank (from which the list of disambiguating connectives is generated), participants are presented with a default list of twelve connectives expressing a variety of relations. 
Based on the connectives chosen in the two steps, the inferred relation sense can be extracted.
For example, the \textsc{Conjunction} reading in Fig.~\ref{fig:fig1} can be expressed by \textit{in addition}, and the \textsc{Result} reading can be expressed by \textit{consequently}. 

The DC method was used to create a crowdsourced corpus of 6,505 discourse-annotated implicit relations, named DiscoGeM \citep{scholman2022DiscoGeM}.
A subset of DiscoGeM is used in the current study (see Section \ref{sec:method}).

\subsubsection{Crowdsourcing DRs by QA method}
\label{sec:crowdsourcing}
\citet{pyatkin2020} proposed to crowdsource discourse relations using QA pairs. They collected a dataset of intra-sentential QA annotations which aim to represent discourse relations by including one of the propositions in the question and the other in the respective answer, with the question prefix (\textit{What is similar to..?, What is an example of..?}) mapping to a relation sense.
Their method was later extended to also work inter-sententially \citep{scholman2022design}. In this work we make use of the extended approach that relates two distinct sentences through a question and answer. The following QA pair, for example, connects the two sentences in Fig.~\ref{fig:fig1} with a \textsc{result} relation.

\ex. \label{QAex} \textbf{What is the result of} Caesar being assassinated by a group of rebellious senators?(S1) - \textit{A new series of civil wars broke out [...]}(S2)

The annotation process consists of the following steps: From two consecutive sentences, annotators are asked to choose a sentence that will be used to formulate a question. The other sentence functions as an answer to that question. Next they start building a question by choosing a question prefix and by completing the question with content from the chosen sentence.

Since it is possible to choose either of the two sentences as question/answer for a specific set of symmetric relations, (i.e. \textbf{What is the reason} a new series of civil wars broke out?), we consider both possible formulations as equivalent. 

The set of possible question prefixes cover all PDTB 3.0 senses (excluding belief and speech-act relations). The direction of the relation sense, e.g. \textit{arg1-as-denier} vs. \textit{arg2-as-denier}, is determined by which of the two sentences is chosen for the question/answer.
While \citet{pyatkin2020} allowed crowdworkers to form multiple QA pairs per instance, i.e. annotate more than one discourse sense per relation, we decided to limit the task to 1 sense per relation per worker. We took this decision in order for the QA method to be more comparable to the DC method, which also only allows the insertion of a single connective.

\section{Method}\label{sec:method}
\subsection{Data}
We annotated 1,200 inter-sentential discourse relations using both the DC and the QA task design.\footnote{The annotations are available at \url{https://github.com/merelscholman/DiscoGeM}.} 
Of these 1,200 relations, 900 were taken from the DiscoGeM corpus and 300 from the PDTB 3.0.

\paragraph{DiscoGeM relations} 
The 900 DiscoGeM instances that were included in the current study represent different domains: 296 instances were taken from the subset of DiscoGeM relations that were taken from Europarl proceedings \citep[written proceedings of prepared political speech taken from the Europarl corpus;][]{Koehn-Europarl-2005}, 304 instances were taken from the literature subset (narrative text from five English books),\footnote{\textit{Animal Farm} by George Orwell, \textit{Harry Potter and the Philosopher's Stone} by J. K. Rowling, \textit{The Hitchhikers Guide to the Galaxy} by Douglas Adam, \textit{The Great Gatsby} by F.~Scott Fitzgerald and \textit{The Hobbit} by J. R. R. Tolkien} and 300 instances from the Wikipedia subset of DiscoGeM (informative text, taken from the summaries of 30 Wikipedia articles).
These different genres enable a cross-genre comparison. This is necessary, given that prevalence of certain relation types can differ across genres \citep{rehbein2016,scholman2022DiscoGeM,webber2009}. 

These 900 relations were already labeled using the DC method in DiscoGeM; we additionally collect labels using the QA method for the current study.
In addition to crowd-sourced labels using the DC and QA methods, the Wikipedia subset was also annotated by three trained annotators.\footnote{Instances were labeled by two annotators and verified by a third; Cohen's $\kappa$ agreement between the first annotator and the reference label was .82 (88\% agreement), and between the second and the reference label was .96 (97\% agreement). See \citet{scholman2022DiscoGeM} for additional details.} 47\% of these Wikipedia instances were labeled with multiple senses by the expert annotators (i.e., were considered to be ambiguous or express multiple readings).

\paragraph{PDTB relations} 
The PDTB relations were included for the purpose of comparing our annotations with traditional PDTB gold standard annotations.
These instances (all inter-sentential) were selected to represent all relational classes, randomly sampling at most 15 and at least 2 (for classes with less than 15 relation instances we sampled all existing relations) relation instances per class.
The reference labels for the PDTB instances consist of the original PDTB labels annotated as part of the PDTB3 corpus. Only 8\% of these consisted of multiple senses.

\subsection{Crowdworkers}\label{crowd-annotators}
Crowdworkers were recruited via Prolific using a selection approach \citep{scholman2022design}, which has been shown to result in a good trade off between quality and time/monetary efforts for DR annotation. 
Crowdworkers had to meet the following requirements: be native English speakers, reside in UK, Ireland, USA, or Canada, and have obtained at least an undergraduate degree. 

Workers who fulfilled these conditions could participate in an initial recruitment task, for which they were asked to annotate a text with either the DC or QA method and were shown immediate feedback on their performance. Workers with an accuracy $\geq 0.5$ on this task were qualified to participate in further tasks. We hence created a unique set of crowdworkers for each method. 
The DC annotations (collected as part of DiscoGeM) were provided by a final set of 199 selected crowdworkers; QA had a final set of 43 selected crowdworkers.\footnote{The larger set of selected workers in the DC method is because more data was annotated by DC workers as part of the creation of DiscoGeM.}
Quality was monitored throughout the production data collection and qualifications were adjusted according to performance.

Every instance was annotated by 10 workers per method. This number was chosen based on parity with previous research. For example, \newcite{snow2008} show that a sample of 10 crowdsourced annotations per instance yields satisfactory accuracy for various linguistic annotation tasks. \newcite{scholman2017context} found that assigning a new group of 10 annotators to annotate the same instances resulted in a near-perfect replication of the connective insertions in an earlier DC study.

Instances were annotated in batches of 20. For QA, one batch took about 20 minutes to complete, and for DC 7 minutes.
Workers were reimbursed about £2.50 and £1.88 per batch respectively.

\subsection{Inter-annotator agreement}\label{sec:iaa}
We evaluate the two DR annotation methods by the inter-annotator agreement (IAA) between the annotations collected by both methods and IAA with reference annotations collected from trained annotators.

Cohen's kappa \cite{cohen1960} is a metric frequently used to measure inter-annotator agreement (IAA).
For DR annotations, a Cohen's kappa of .7 is considered to reflect good IAA \citep{spooren2010}. However, prior research has shown that agreement on implicit relations is more difficult to reach than on explicit relations: 
\newcite{Kishimoto2018ImprovingCA} report an F1 of .51 on crowdsourced annotations of implicits using a tagset with 7 level-2 labels;
\newcite{zikanova2019explicit} report $\kappa$=.47 (58\%) on expert annotations of implicits using a tagset with 23 level-2 labels; and \newcite{demberg2019compatible} find that PDTB and RST-DT annotators agree on the relation sense on 37\% of implicit relations.
Cohen's kappa is primarily used for comparison between single labels and the IAAs reported in these works are also based on single aggregated labels.

However, we also want to compare the obtained 10 annotations per instance with our reference labels which also contain multiple labels. 
 The comparison becomes less straightforward when there are multiple labels because the chance of agreement is inflated and partial agreement should be treated differently.  We thus measure the IAA between multiple labels in terms of both full and partial agreement rates, as well as the multi-label kappa metric proposed by \citet{marchal2022establishing}.  This metric adjusts the multi-label agreements with bootstrapped expected agreement.
 We consider all the labels annotated by the crowdworkers in each instance, excluding minority labels with only one vote\footnote{We assumed there were $10$ votes per item and removed labels with less than 20\% of votes, even though in rare cases there could be $9$ or $11$ votes. On average, the removed labels represent $24.8\%$ of the votes per item.}.

In addition, we compare the distributions of the crowdsourced labels using the Jensen-Shannon divergence (JSD) following existing works \cite{erk2009graded,nie2020can,zhang2021learning}. 
Similarly, minority labels with only one vote are excluded. Since distributions are not available in the reference labels, when comparing with the reference labels, we evaluate by the JSD based on the flattened distributions of the labels, which means we replace the original distribution of the votes with an even distribution of the labels that have been voted by more than one annotator. We call this version JSD\_{flat}.

\begin{table*}[ht]
    \centering\small
\begin{tabular}{@{}llllllll@{}}
\hline
 & Europarl   & Novel     & Wiki.     & PDTB      
 & all  \\
 
Item counts  & 296        & 304       & 300           & 302       
& 1202 \\ \hline
$QA$ sub-labels/item   & 2.13     & 2.21     & 2.26  & 2.45    
& 2.21   \\  
$DC$ sub-labels/item   & 2.37      & 2.00     & 2.09  & 2.21       
& 2.17    \\ \hline\hline
 
full/+partial   agreement   & .051/.841  & .092/.865 &  .060/.920     & .050/.884      
& .063/.878     \\

multi-label kappa & .813  & .842 & .903 & .868 
& .857\\
JSD & .505 & .492 & .482 & .510 
& .497 \\
\hline
\end{tabular}
\caption{Comparison between the labels obtained by DC vs. QA. Full (or +partial) agreement means all (or at least one sub-label) match(es).
Multi-label kappa is adapted from \citet{marchal2022establishing}. 
JSD 
is calculated based on the actual 
distributions of the crowdsourced sub-labels, excluding labels with only one vote (smaller values are better). 
}
\label{tab:IAA_methods}
\end{table*}

As a third perspective on IAA we report agreement among annotators on an item annotated with QA/DC.
 Following previous work \cite{nie2020can}, we use entropy of the soft labels to quantify the uncertainty of the crowd annotation.   Here labels with only one vote are also included as they contribute to the annotation uncertainty.   When calculating the entropy, we use a logarithmic base of $n=29$, where $n$ is the number of possible labels. A lower entropy value suggests that the annotators agree with each other more and the annotated label is more certain.  
As discussed in Sec.~\ref{sec:introduction}, the source of disagreement in annotations could come from the items, the annotators and the methodology. High entropy across multiple annotations of a specific item within the same annotation task suggests that the item is ambiguous.

\section{Results}\label{sec:results}

We first compare the IAA between the two crowdsourced annotations, then we discuss IAA between DC/QA and the reference annotations, and lastly we perform an analysis based on annotation uncertainty. Here, "sub-labels" of an instance means all relations that have received more than one annotation; and "label distribution" is the distribution of the votes of the sub-labels.

\subsection{IAA between the methods}
Tab.~\ref{tab:IAA_methods} shows that both methods yield more than two sub-labels per instance after excluding minority labels with only one vote. This supports the idea that \textbf{multi-sense annotations better capture the fact that often more than one sense can hold implicitly between two discourse arguments}. 

Tab.~\ref{tab:IAA_methods} also presents the IAA between the labels crowdsourced with QA and DC per domain. 
The agreement between the two methods is good: the labels assigned by the two methods (or at least one of the sub-labels in case of a multi-label annotation) match
for about $88\%$ of the items. 
This speaks for the fact that \textbf{both methods are valid}, as similar sets of labels are produced. 

The full agreement scores, however, are very low. This is expected, as the chance to match on all sub-labels is also very low compared to a single-label setting. The multi-label kappa -- which takes chance agreement of multiple labels into account--, and JSD -- which compares the distributions of the multiple labels--, are hence more suitable. We note that the PDTB gold annotation that we use for evaluation does not assign multiple relations systematically and has a low rate of double labels. This explains why the PDTB subsets have a high partial agreement while the JSD ends up being worst.

\subsection{IAA between crowdsourced and reference labels}

\begin{table}
\centering
\begin{tabular}{@{}lll@{}}

\hline
   & Wiki.     & PDTB       \\ 
       
Item counts   & 300   & 302  \\ \hline

Ref. sub-labels/item    & 1.54  & 1.08  \\

\hline\hline
QA: sub-labels/item   &  2.26   &  2.45    \\ 
full/+partial agreement   & .133/.887   & .070/.487     \\ 

multi-label kappa & .857  & .449 \\
JSD$_{flat}$    
&  .468 & .643   \\
\hline
DC: sub-labels/item  &   2.09  &  2.21  \\                             
full/+partial  agreement   & .110/.853   & .103/.569  \\

multi-label kappa & .817  & .524 \\ 
JSD$_{flat}$    
& .483  & .606 \\
\hline

\hline          
\end{tabular}
\caption{Comparison against gold labels for the QA or DC methods.
Since the distribution of the reference sub-labels is not available, JSD\_{flat} is calculated between uniform distributions of the sub-labels.
\label{tab:IAA_ref}}
\end{table}

Table~\ref{tab:IAA_ref} compares the labels crowdsourced by each method and the reference labels, which are available for the Wikipedia and PDTB subsets. It can be observed that both methods achieve higher full agreements with the reference labels than with each other on both domains. This indicates that the \textbf{two methods are complementary}, with each method better capturing different sense types.
In particular, the QA method tends to show higher agreement with the reference for Wikipedia items,  while the DC annotations show higher agreement with the reference for PDTB items.
This can possibly be \textbf{attributed to the development of the methodologies}: the DC method was originally developed by testing on data from the PDTB in \citet{yung2019}, whereas the QA method was developed by testing on data from Wikipedia and Wikinews in \citet{pyatkin2020}.

\begin{figure}[htpb]
    \centering 
\includegraphics[width=0.45\textwidth, trim= 0cm 0cm 0cm 0cm, clip=TRUE]{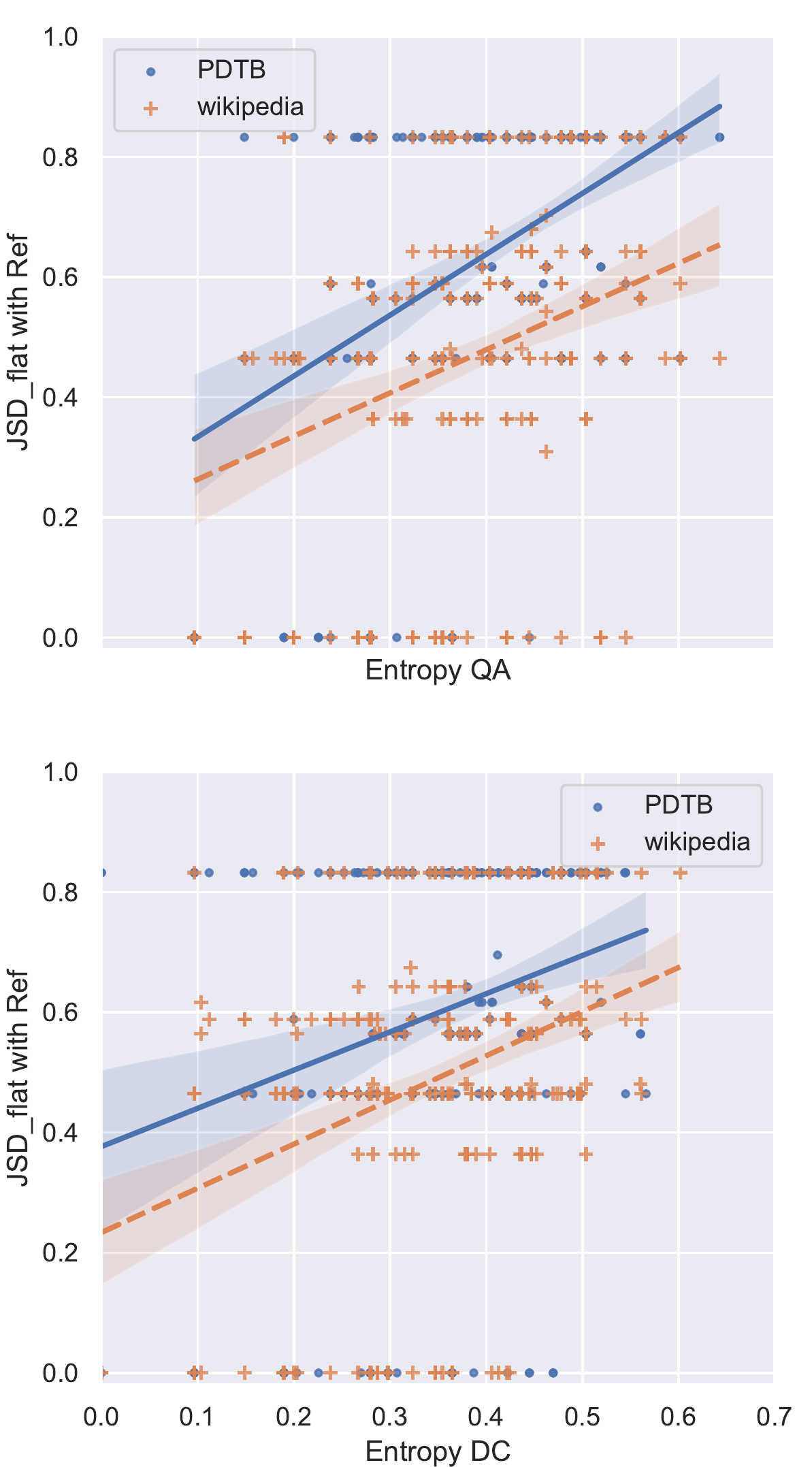}
\caption{Correlation between the entropy of the annotations and the JSD$_{flat}$ 
between the crowdsourced labels and reference. }
\label{Fig:ent_vs_ref_JSD}
\end{figure}

\subsection{Annotation uncertainty}

\begin{table}[ht]
    \centering
\begin{tabular}{lllll}
\hline
   & Europarl & Wikipedia & Novel & PDTB \\\hline\hline
QA & 0.40     & 0.38      & 0.38  & 0.41 \\\hline
DC & 0.37     & 0.34      & 0.35  & 0.36 \\ 
\hline
\end{tabular}
    \caption{Average entropy of the label distributions (10 annotations per relation) for QA/DC, split by domain.} 
    \label{tab:entropy}
\end{table}

Table~\ref{tab:entropy} compares the average entropy of the soft labels collected by both methods. It can be observed that the uncertainty among the labels chosen by the crowdworkers is similar across domains but always slightly lower for DC. We further look at the correlation between annotation uncertainty and cross-method agreement, and find that agreement between methods is substantially higher for those instances where within-method entropy was low. Similarly, we find that agreement between crowdsourced annotations and gold labels is highest for those relations, where little entropy was found in crowdsourcing.

Next, we want to check if the item effect is similar across different methods and domains. Figure \ref{Fig:ent_vs_ref_JSD} in the Appendix shows the correlation between the annotation entropy and the agreement with the reference of each item, of each method for the Wikipedia / PDTB subsets. 
It illustrates that annotations of both methods diverge with the reference more as the uncertainty of the annotation increases.  While the effect of uncertainty is similar across methods on the Wikipedia subset, the quality of the QA annotations depends more on the uncertainty compared to the DC annotations on the PDTB subset.  This means that method bias also exists on the level of annotation uncertainty and should be taken into account when, for example, entropy is used as a criterion to select reliable annotations.

\section{Sources of the method bias}
\label{sec:method_bias_sources}
In this section, we analyze method bias in terms of the sense labels collected by each method. We also examine the potential limitations of the methods which could have contributed to the bias and demonstrate how we can utilize information on method bias to crowdsource more reliable labels. Lastly, we provide a cross-domain analysis.

\begin{figure}[ht]
    \centering
    \includegraphics[width=0.5\textwidth]{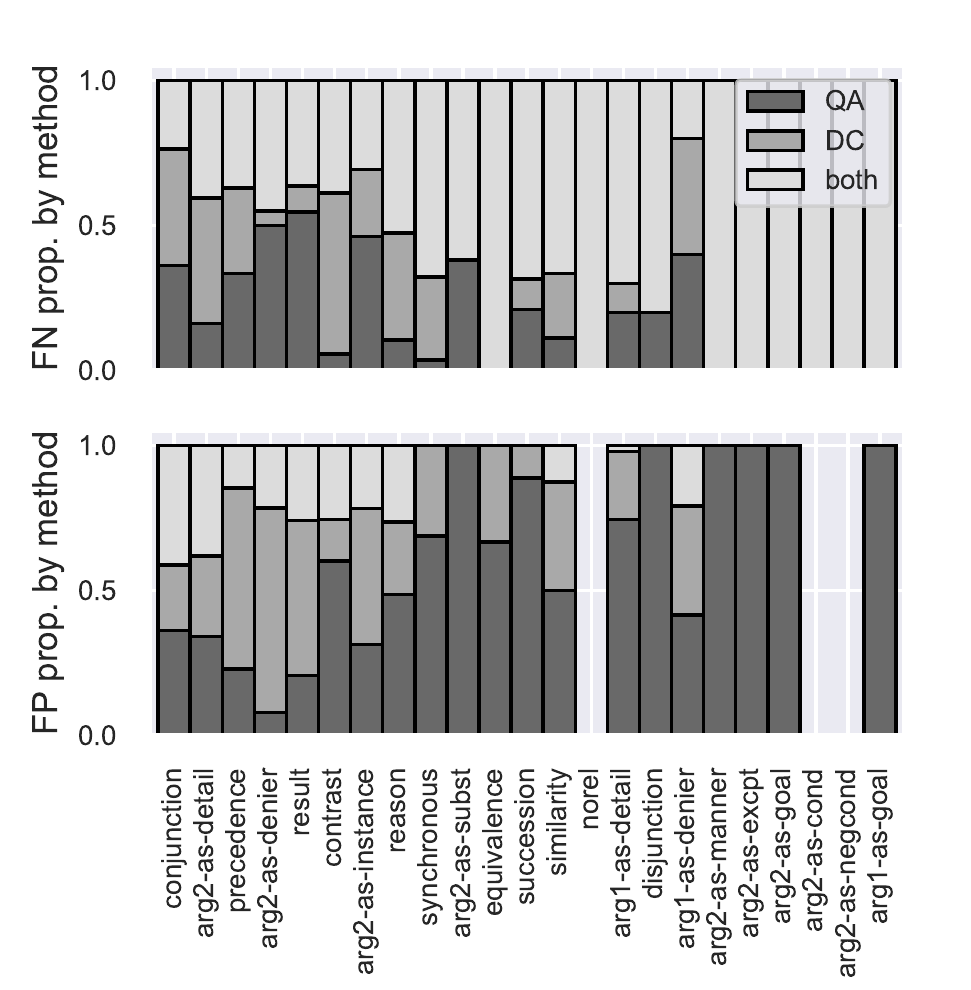}
    \caption{Distribution of the annotation errors by method. Labels annotated by at least 2 workers are compared against the reference labels of the Wikipedia and PDTB items. The relation types are arranged in descending order of the "ref. sub-label counts"
    }
    \label{fig:error_dist}
\end{figure}

\begin{table}[htpb]
    \centering \small
    \begin{tabular}{l|llll}
    \hline
    label & FN$_{QA}$ & FN$_{DC}$ & FP$_{QA}$ & FP$_{DC}$\\
    \hline
conjunction & 43 & 46 & 203 & 167 \\
arg2-as-detail & 42 & 62 & 167 & 152 \\
precedence & 19 & 18 & 18 & 37 \\
arg2-as-denier & 38 & 20 & 15 & 47 \\
result & 10 & 5 & 110 & 187 \\
contrast & 8 & 17 & 84 & 39 \\
arg2-as-instance & 10 & 7 & 44 & 57 \\
reason & 12 & 17 & 54 & 37 \\
synchronous & 20 & 27 & 11 & 5 \\
arg2-as-subst & 21 & 13 & 1 & 0 \\
equivalence & 22 & 22 & 2 & 1 \\
succession & 17 & 15 & 24 & 3 \\
similarity & 7 & 8 & 15 & 12 \\
norel & 12 & 12 & 0 & 0 \\
arg1-as-detail & 9 & 8 & 39 & 13 \\
disjunction & 5 & 4 & 10 & 0 \\
arg1-as-denier & 3 & 3 & 33 & 31 \\
arg2-as-manner & 2 & 2 & 9 & 0 \\
arg2-as-excpt & 2 & 2 & 1 & 0 \\
arg2-as-goal & 1 & 1 & 5 & 0 \\
arg2-as-cond & 1 & 1 & 0 & 0 \\
arg2-as-negcond & 1 & 1 & 0 & 0 \\
arg1-as-goal & 1 & 1 & 3 & 0 \\
\hline
    \end{tabular}
    \caption{FN and FP counts of each method grouped by the reference sub-labels}
    \label{tab:FN_FP}
\end{table}

Table \ref{tab:heatmap_subset} presents the confusion matrix of the labels collected by both methods for the most frequent level-2 relations. 
 Figure \ref{fig:error_dist} and Table \ref{tab:FN_FP} in the Appendix shows the distribution of the true and false positives of the sub-labels. These results show that both methods are biased towards certain DRs. The source of these biases can be categorized into two types, which we will detail in the following subsections.

\begin{table}[htpb]
\small
\setlength{\tabcolsep}{4pt}
\begin{tabular}{@{}l|@{}rrrrrrrr|r@{}} \hline
DC    \hspace{1em} \sid{QA}  & \sid{Synch.} & \sid{Asynch.} & \sid{Cause} & \sid{Concession} & \sid{Contrast} & \sid{Conjunction} & \sid{Detail} & \sid{Instant.} & \sid{all DC}  \\ \hline
\multicolumn{1}{l|}{Synchr.} & \multicolumn{1}{r|}{1}  & 1                       & 1                        &                        &                         & 2                        & 1                       &                         & 6 \\ \cline{2-3}
Asynchr.                     & \multicolumn{1}{r|}{2} & \multicolumn{1}{r|}{\cellcolor{heat6}{45}} & \cellcolor{heat8}{17}                       &                        & 7                       & \cellcolor{heat8}{18}                       & \cellcolor{heat10}{7}                       & 2                       & 98 \\ \cline{3-4}
Cause                            &                        & \multicolumn{1}{r|}{\cellcolor{heat8}{13}} & \multicolumn{1}{r|}{\cellcolor{heat1}{184}} & 5                      & \cellcolor{heat10}{8}                       & \cellcolor{heat4}{78}                       & \cellcolor{heat7}{36}                      & \cellcolor{heat8}{13}                      & 337 \\ \cline{4-5}
Concess.                       & 2                      & 3                       & \multicolumn{1}{r|}{\cellcolor{heat8}{13}}  & \multicolumn{1}{r|}{\cellcolor{heat10}{14}} & \cellcolor{heat8}{22}                      & \cellcolor{heat8}{18}                       & \cellcolor{heat8}{22}                      & 1                       & 95 \\ \cline{5-6}
Contrast                         &                1        &                         & \cellcolor{heat10}{7}                        & \multicolumn{1}{r|}{1}  & \multicolumn{1}{r|}{\cellcolor{heat8}{21}} & 3                        &          4               &                         & 37 \\ \cline{6-7}
Conjunct.                      & 4                      & \cellcolor{heat8}{12}                      & {\cellcolor{heat6}37}                       & 1                      & \multicolumn{1}{r|}{\cellcolor{heat10}{12}}  & \multicolumn{1}{r|}{\cellcolor{heat1}{184}} & \cellcolor{heat4}{67}                      & 6                       & 323 \\ \cline{7-8}
Detail                  & 2                      & 5                       & \cellcolor{heat6}{40}                       & 2                      & 4                       & \multicolumn{1}{r|}{\cellcolor{heat5}{52}}  & \multicolumn{1}{r|}{\cellcolor{heat4}{93}} & \cellcolor{heat8}{16}                      & 214 \\ \cline{8-9} 
Instant.                    &                        &                         & 2                        & 1                      & 2                       & \cellcolor{heat10}{9}                        & \multicolumn{1}{r|}{\cellcolor{heat10}{10}}  & \multicolumn{1}{r|}{\cellcolor{heat8}{23}} & 47 \\ \cline{9-9} 
   \hline
all QA & 12  & 79 & 301  & 24   & 76  & 364 & 240  & 61 \\ %\cline{9-9} 
\hline
\end{tabular}
\caption{Confusion matrix for the most frequent level-2 sublabels which were annotated by at least 2 workers per relation; values are represented as colors.}
\label{tab:heatmap_subset}
\end{table}

\subsection{Limitation of natural language for annotation}
\label{sec:NL_limitation}

There are limitations of representing DRs in natural languages using both QA and DC.  For example, the QA method confuses workers when the question phrase contains a connective:\footnote{The examples are presented in the following format: \textit{italics} = argument 1; \textit{bolded} = argument 2; plain = contexts.}

\ex. \label{eg:succession}
    \textit{``Little tyke, "chortled Mr. Dursley as he left the house.} \textbf{He got into his car and backed out of number four's drive.}
    [QA:\textsc{succession, precedence}, DC:\textsc{conjunction, precedence}]
\normalsize

In the above example, the majority of the workers formed the question ``\textbf{After what} \textit{he left the house?}, which was likely a confusion with ``\textbf{What did he do after} \textit{he left the house?"}. This could explain the frequent confusion between \textsc{precedence} and \textsc{succession} by QA, resulting in the frequent FPs of \textsc{succession} (Fig.~\ref{fig:error_dist}).\footnote{Similarly, the question \textit{``Despite what ... ?"} is easily confused with \textit{``despite..."}, which could explain the frequent FP of \textit{arg1-as-denier} by the QA method.} 

\medskip
For DC, rare relations which lack a frequently used connective are harder to annotate, for ex.: 
\ex. \label{eg:arg1instance} 
     \textit{He had made an arrangement with one of the cockerels to call him in the mornings half an hour earlier than anyone else, and would put in some volunteer labour at whatever seemed to be most needed, before the regular day's work began.} \textbf{His answer to every problem, every setback, was ``I will work harder!" - which he had adopted as his personal motto.}
     [QA:\textsc{arg1-as-instance}; DC:\textsc{result}]
\normalsize

It is difficult to use the DC method to annotate the \textsc{arg1-as-instance} relation due to a lack of typical, specific and context independent connective phrases that mark these rare relations, such as "\textit{this is an example of ...}".  By contrast, the QA method allows workers to make a question and answer pair in the reverse direction, with S1 being the answer to S2, using the same question words, e.g. \textit{\textbf{What is an example} of the fact that his answer to every problem [...] was ``I will work harder!"?}. This allows workers to label rarer relation types that were not even uncovered by trained annotators.

Many common DCs are ambiguous, such as \textit{but} and \textit{and}, and can be hard to disambiguate. To address this, the DC method provides workers with unambiguous connectives in the second step. However, these unambiguous connectives are often relatively uncommon and come with different syntactic constraints, depending on whether they are coordinating or subordinating conjunctions or discourse adverbials. Hence, they do not fit in all contexts. Additionally, some of the unambiguous connectives sound very ``heavy'' and would not be used naturally in a given sentence. For example, \textit{however} is often inserted in the first step, but it can mark multiple relations and is disambiguated in the second step by the choice among \textit{on the contrary} for \textsc{contrast}, \textit{despite} for \textsc{arg1-as-denier} and \textit{despite this} for \textsc{arg2-as-denier}. \textit{Despite this} was chosen frequently since it can be applied to most contexts. This explains the DC method's bias towards \textit{arg2-as-denier} against \textit{contrast} (Figure \ref{fig:error_dist}: most FPs of \textit{arg2-as-denier} and most FNs of \textit{contrast} come from DC).

While the QA method also requires workers to select from a set of question starts, which also contain infrequent expressions (such as \textit{Unless what..?}), workers are allowed to edit the text to improve the wordings of the questions.  This helps reduce the effect of bias towards more frequent question prefixes and makes crowdworkers doing the QA task more likely to choose infrequent relation senses than those doing the DC task.

\subsection{Guideline underspecification}
\label{sec:unclear_guidelines}
\citet{jiang2022investigating} report that some disagreements in NLI tasks come from the loose definition of certain aspects of the task. We found that both QA and DC also \textbf{do not give clear enough instructions in terms of argument spans}. The DRs are annotated at the boundary of two consecutive sentences but both methods do not limit workers to annotate DRs that span exactly the two sentences.

More specifically, the QA method allows the crowdworkers to form questions by copying spans from one of the sentences. While this makes sure that the relation lies locally between two consecutive sentences, it also sometimes happens that workers highlight partial spans and annotate relations that span over parts of the sentences. For ex.:

\ex. \label{eg:europarl} 
    \textit{I agree with Mr Pirker, and it is probably the only thing I will agree with him on if we do vote on the Ludford report.} \textbf{It is going to be an interesting vote.}
    [QA:\textsc{arg2-as-detail,reason}; DC:\textsc{conjunction,result}]

In Ex.~\ref{eg:europarl}, workers constructed the question ``\textbf{What provides more details} \textit{on the vote on the Ludford report?}".
This is similar to the instructions in PDTB 2.0 and 3.0's annotation manuals, specifying that annotators should take minimal spans which don't have to span the entire sentence.
Other relations should be inferred when the argument span is expanded to the whole sentence, for example a \textsc{result} relation reflecting that there is little agreement, which will make the vote interesting. 

Often, a sentence can be interpreted as the elaboration of certain entities in the previous sentence. This could explain why \textsc{Arg1/2-as-detail} tends to be overlabelled by QA.  Fig.~\ref{fig:error_dist} shows that the QA has more than twice as many FP counts for \textsc{arg2-as-detail} compared to DC -- the contrast is even bigger for \textsc{arg1-as-detail}. 
Yet it is not trivial to filter out such questions that only refer to a part of the sentence, because in some cases, the highlighted entity does represent the whole argument span.\footnote{Such as ``\textit{a few final comments}" in this example:
\textit{Ladies and gentlemen, I would like to make a few final comments.} \textbf{This is not about the implementation of the habitats directive.}}
Clearer instructions in the guidelines are desirable.

Similarly, DC does not limit workers to annotate relations between the two sentences, consider:
\ex. \label{eg:semiconductor} 

    \textit{When two differently-doped regions exist in the same crystal, a semiconductor junction is created.} \textbf{The behavior of charge carriers, which include electrons, ions and electron holes, at these junctions is the basis of diodes, transistors and all modern electronics.}
    [Ref:\textsc{arg2-as-detail}; QA:\textsc{arg2-as-detail, conjunction}; DC:\textsc{conjunction, result}]   
\normalsize

In this example, many people inserted \textit{as a result}, which naturally marks the intra-sentence relation (\textit{...is created as a result.})
Many relations are potentially spuriously labelled as \textsc{Result}, which are frequent between larger chunks of texts. 
Tab.~\ref{tab:heatmap_subset} shows that the most frequent confusion is between DC's \textsc{cause} and QA's \textsc{conjunction}.\footnote{A chi-squared test confirms that the observed distribution is significantly different from what could be expected based on chance disagreement.}
Within the level-2 \textsc{cause} relation sense, it is the level-3 \textsc{result} relation that turns out to be the main contributor to the observed bias. 
Fig.~\ref{fig:error_dist} also shows that most FPs of \textsc{result} come from the DC method.

\subsection{Aggregating DR annotations based on method bias}
The qualitative analysis above provides insights on certain method biases observed in the label distributions, such as QA's bias towards \textsc{arg1/2-as detail} and \textsc{succession} and DC's bias towards \textsc{concession} and \textsc{result}. Being aware of these biases would allow to combine the methods: after first labelling all instances with the more cost-effective DC method, \textsc{result} relations, which we know tend to be overlabelled by the DC method, could be re-annotated using the QA method. We simulate this for our data and find that this would increase the partial agreement from 0.853 to 0.913 for wikipedia and from 0.569 to 0.596 for PDTB.

\section {Analysis by Genre}
\label{sec:genre}

\begin{figure*}[!ht]
    \centering % <-- added
\includegraphics[width=16cm, trim= 0cm 0.2cm 0cm 0cm, clip=TRUE]{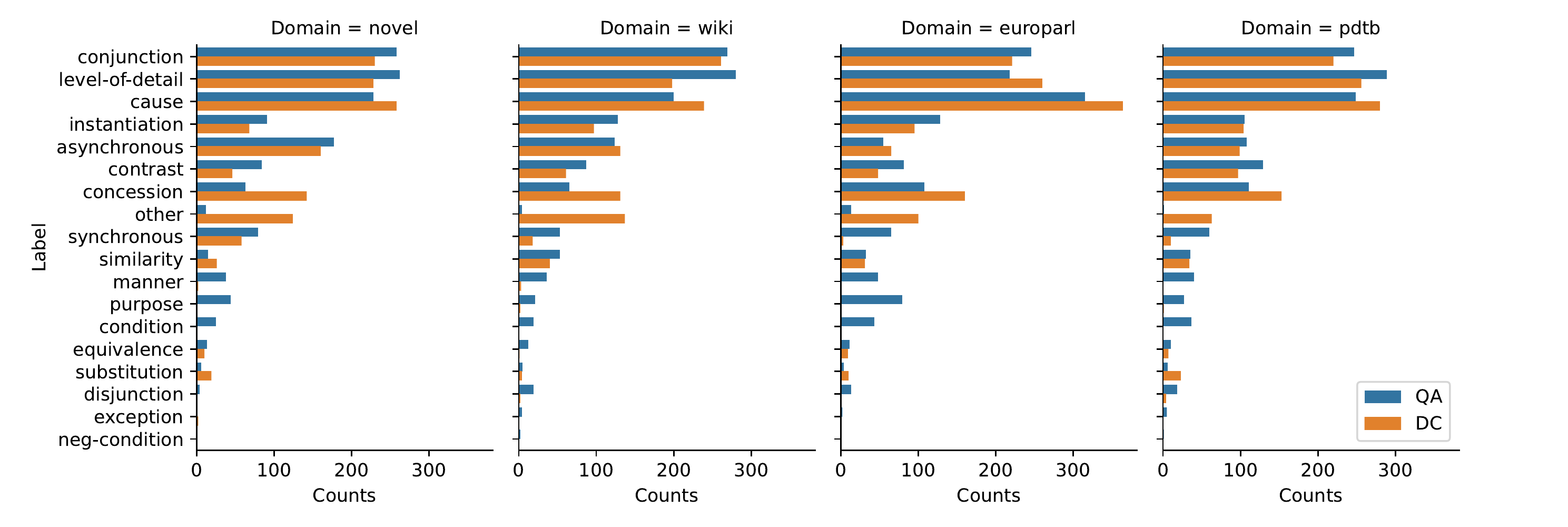}
\caption{Level-2 sublabel counts of all the annotated labels of both methods, split by domain. }
\label{Fig:domains}
\end{figure*}
For each of the four genres (novel, wikipedia, europarl and wsj) we have \textasciitilde300 implicit DRs annotated by both DC and QA.
 \citet{scholman2022DiscoGeM} showed, based on the DC method, that in DiscoGeM, \textsc{conjunction} is prevalent in the Wikipedia domain, \textsc{Precedence} in Literature and \textsc{Result} in Europarl. The QA annotations replicate this finding, as displayed in Fig.~\ref{Fig:domains}.

It appears more difficult to obtain agreement with the majority labels in Europarl than in other genres, which is reflected in the average entropy (see Table \ref{tab:entropy}) of the distributions for each genre, where DC has the highest entropy in the Europarl domain and QA the second highest (after PDTB).  
Table \ref{tab:IAA_methods} confirms these findings, showing that the agreement between the two methods is highest for Wikipedia and lowest for Europarl. 

In the latter domain, the DC method results in more \textsc{causal} relations: 36\% of the \textsc{conjunctions} labelled by QA are labelled as \textsc{result} in DC.\footnote{This appeared to be distributed over many annotators and is thus a true method bias.}
Manual inspection of these DC annotations reveals that workers chose \textit{considering this} frequently only in the Europarl subset.  This connective phrase is typically used to mark a pragmatic result relation, where the result reading comes from the belief of the speaker (Ex.~\ref{eg:europarl}). This type of relation is expected to be more frequent in speech and argumentative contexts and is labelled as \textsc{result-belief} in PDTB3. QA does not have a question prefix available that could capture \textsc{result-belief} senses. The \textsc{result} labels obtained by DC are therefore a better fit with the PDTB3 framework than QA's \textsc{conjunctions}.
\textsc{Concession} is generally more prevalent with the DC method, especially in Europarl, with 9\% compared to 3\% for QA. \textsc{Contrast}, on the other hand, seems to be favored by the QA method, of which most (6\%) \textsc{contrast} relations are found in Wikipedia, compared to 3\% for DC. 
Figure \ref{Fig:domains} also highlights that for the QA approach, annotators tend to choose a wider variety of senses which are rarely ever annotated by DC, such as \textsc{purpose}, \textsc{condition} and \textsc{manner}.

We conclude that encyclopedic and literary texts are the most suitable to be annotated using either DC or QA, as they show higher inter-method agreement (and for Wikipedia also higher agreement with gold). Spoken-language and argumentative domains on the other hand are trickier to annotate as they contain more pragmatic readings of the relations.

\section{Case Studies: Effect of task design on DR classification models}
\label{sec:models}
Analysis of the crowdsourced annotations reveals that the two methods have different biases and 
different correlations with domains and the
style (and possibly function) of the language used in the domains.
We now investigate the effect of task design bias on automatic prediction of implicit discourse relations. Specifically, we carry out two case studies to demonstrate the effect that task design and the resulting label distributions have on discourse parsing models.

\paragraph{Task and setup}
We formulate the task of predicting implicit discourse relations as follows. The input to the model are two sequences $S_1$ and $S_2$, which represent the arguments of a discourse relation. 
The targets are PDTB 3.0 sense types (including level-3).
This model architecture is similar to the model for implicit DR prediction by \citet{shi2019}.
We experiment with two different losses and targets: a cross-entropy loss where the target is a single majority label and a soft cross-entropy loss where the target is a probability distribution over the annotated labels. Using the 10 annotations per instance we obtain label distributions for each relation, which we use as soft targets. Training with a soft loss has been shown to improve generalization in vision and NLP tasks \cite{peterson2019human, uma2020case}. As suggested in \citet{uma2020case}, we normalize the sense-distribution over the 30 possible labels\footnote{precedence, arg2-as-detail, conjunction, result, arg1-as-detail, arg2-as-denier, contrast, arg1-as-denier, synchronous, reason, arg2-as-instance, arg2-as-cond, arg2-as-subst, similarity, disjunction, succession, arg1-as-goal, arg1-as-instance, arg2-as-goal, arg2-as-manner, arg1-as-manner, equivalence, arg2-as-excpt, arg1-as-excpt, arg1-as-cond, differentcon, norel, arg1-as-negcond, arg2-as-negcond, arg1-as-subst
} with a softmax. 

Assuming one has a relation with the following annotations: 4 \textsc{result}, 3 \textsc{conjunction}, 2 \textsc{succession}, 1 \textsc{arg1-as-detail}. For the hard loss, the target would be the majority label: \textsc{result}. For the soft loss we normalize the counts (every label with no annotation has a count of 0) using a softmax, for a smoother distribution without zeros.

We fine-tune DeBERTa (deberta-base) \cite{he2020deberta} in a sequence classification setup using the huggingface checkpoint \cite{wolf2019huggingface}. The model trains for 30 epochs with early stopping and a batch size of 8. 

\paragraph{Data}
In addition to the 1,200 instances we analyzed in the current contribution, we additionally use all annotations from DiscoGeM as training data. DiscoGeM, which was annotated with the DC method, adds 2756 \textit{Novel} relations, 2504 \textit{Europarl} relations and 345 \textit{Wikipedia} relations. We formulate different setups for the case studies.

\subsection{Case 1: incorporating data from different task designs}
The purpose of this study is to see if a model trained on data crowdsourced by DC/QA methods can generalize to traditionally annotated test sets.
We thus test on the 300 Wikipedia relations annotated by experts (\textit{Wiki gold}), all implicit relations from the test set of PDTB 3.0 (\textit{PDTB test}) and the implicit relations of the English test set of TED-MDB \cite{zeyrek2020ted}. 
For training data, we either use 1) all of the DiscoGeM annotations (\textbf{Only DC}); or 2) 1200 QA annotations from all four domains, plus 5,605 DC annotations from the rest of DiscoGeM (\textbf{Intersection}, $\cap$); or 3) 1200 annotations which combine the label counts (e.g. 20 counts instead of 10) of QA and DC, plus 5,605 DC annotations from the rest of DiscoGeM (\textbf{Union}, $\cup$). We hypothesize that this union will lead to improved results due to the annotation distribution coming from a bigger sample.
When testing on \textit{Wiki gold}, the corresponding subset of \textit{Wikipedia} relations are removed from the training data.  
We randomly sampled 30 relations for dev.

\paragraph{Results}

\begin{table}[t]
\small
\begin{tabular}{llll}
\hline
                                & PDTB$_{test}$ & Wiki$_{gold}$ & TED-M.\\ \hline\hline
DC & 0.34$\dagger$      & 0.65$\dagger$    & 0.36\\
DC Soft & 0.29*$\dagger$      & \textbf{0.70}*$\dagger$    & 0.34*\\
QA+DC$_\cap$       & 0.34$\star$       &   0.67   & 0.37 \\
QA+DC$_\cap$ Soft   & 0.38*$\star$       &  0.66*    & 0.31* \\
QA+DC$_\cup$  & 0.35$\spadesuit$ & 0.49$\spadesuit$ & 0.36$\spadesuit$\\
QA+DC$_\cup$ Soft& \textbf{0.41}*$\spadesuit$ & 0.67*$\spadesuit$& \textbf{0.43}*$\spadesuit$ \\
\hline
\end{tabular}
    \caption{Accuracy of model (with soft vs. hard loss) prediction on gold labels. The model is trained either on DC data (DC), an intersection of DC and QA ($\cap$) or the union of DC and QA ($\cup$). Same symbol in a column indicates a statistically significant (McNemar test) difference in cross-model results.} 
    \label{tab:mix}
\end{table}

Table.~\ref{tab:mix} shows how the model generalizes to traditionally annotated data. 
On the PDTB and the Wikipedia test set, the model with a soft loss generally performs better than the hard loss model. TED-MDB on the other hand only contains a single label per relation and training with a distributional loss is therefore less beneficial.
Mixing DC and QA data only improves in the soft case for PDTB.
The merging of the respective method label counts, on the other hand, leads to the best model performance on both PDTB and TED-MDB. On Wikipedia the best performance is obtained when training on soft DC-only distributions. Looking at the label-specific differences in performance, we observe that improvement on the Wikipedia test set mainly comes from better precision and recall when predicting \textsc{arg2-as-detail}, while on PDTB QA+DC$_{\cap}$ Soft is better at predicting \textsc{conjunction}.

We conclude that training on data that comes from different task designs does not hurt performance, and even slightly improves performance when using majority vote labels. When training with a distribution, the union setup ($\cup$) seems to work best.

\begin{table*}[t]
\small
\definecolor{grey}{gray}{0.8}.
\begin{tabular}{ll|lll|lll|lll|lll}
\hline
      &    & \multicolumn{3}{l}{\begin{tabular}[c]{@{}l@{}}All except Wiki DC\end{tabular}}                                                                              & \multicolumn{3}{l}{\begin{tabular}[c]{@{}l@{}}All except EP DC\end{tabular}}                                                                                & \multicolumn{3}{l}{\begin{tabular}[c]{@{}l@{}}All except Novel DC\end{tabular}}                                                                             & \multicolumn{3}{l}{All}                                                                                                                                      \\ \hline\hline
      &    & \begin{tabular}[c]{@{}l@{}}hard\\ acc\end{tabular} & \begin{tabular}[c]{@{}l@{}}soft\\ acc\end{tabular} & \begin{tabular}[c]{@{}l@{}}soft\\ JSD\end{tabular} & \begin{tabular}[c]{@{}l@{}}hard\\ acc\end{tabular} & \begin{tabular}[c]{@{}l@{}}soft\\ acc\end{tabular} & \begin{tabular}[c]{@{}l@{}}soft\\ JSD\end{tabular} & \begin{tabular}[c]{@{}l@{}}hard\\ acc\end{tabular} & \begin{tabular}[c]{@{}l@{}}soft\\ acc\end{tabular} & \begin{tabular}[c]{@{}l@{}}soft\\ JSD\end{tabular} & \begin{tabular}[c]{@{}l@{}}hard\\ acc\end{tabular} & \begin{tabular}[c]{@{}l@{}}soft\\ acc\end{tabular} & \begin{tabular}[c]{@{}l@{}}soft\\ JSD\end{tabular} \\
Wiki  & QA &  \cellcolor{grey}{0.51*} & \cellcolor{grey}{\textbf{ 0.53*}} & \cellcolor{grey}{0.38}  &\cellcolor{grey}{ 0.51}   & \cellcolor{grey}{0.50}  & \cellcolor{grey}{0.37}   & \cellcolor{grey}{0.47} & \cellcolor{grey}{0.53*}  & \cellcolor{grey}{0.38}   & \cellcolor{grey}{0.50*}                                                & \cellcolor{grey}{0.51}  & \cellcolor{grey}{0.38}                                            \\
      & DC & 0.55*                                                & 0.59*                                                & 0.36                                             & 0.60                                                & \textbf{0.62}                                                & 0.33                                             & 0.56                                                & 0.61*                                                & 0.35                                             & 0.58*                                                & 0.61                                                & 0.35                                             \\
EP    & QA & \cellcolor{grey}0.30  & \cellcolor{grey}0.35   & \cellcolor{grey}0.43                                             & \cellcolor{grey}0.29   & \cellcolor{grey}0.33                                                &\cellcolor{grey}0.42                                             & \cellcolor{grey}0.29  & \cellcolor{grey}\textbf{0.38}   &\cellcolor{grey}0.44                                             & \cellcolor{grey}0.29                                                & \cellcolor{grey}0.34                                                & \cellcolor{grey}0.44                                             \\
      & DC & 0.50                                                & \textbf{0.53}                                                & 0.34                                             & 0.41                                                & 0.47                                                & 0.37                                             & 0.49                                                & 0.49                                                & 0.36                                             & 0.47                                                & 0.52                                                & 0.36                                             \\
Novel & QA & \cellcolor{grey}0.44*    & \cellcolor{grey}\textbf{0.47}                                                &\cellcolor{grey}0.40                                             &\cellcolor{grey} 0.41                                                &\cellcolor{grey} 0.45                                                & \cellcolor{grey}0.41     & \cellcolor{grey}0.40                                                & \cellcolor{grey}0.39                                                & \cellcolor{grey}0.44                                             & \cellcolor{grey}0.39                                                & \cellcolor{grey}0.45                                                & \cellcolor{grey}0.41                                             \\
      & DC & 0.52*                                                & \textbf{0.58}                                                & 0.33                                             & 0.53                                                & 0.56                                                & 0.34                                             & 0.56                                                & 0.51                                                & 0.39                                             & 0.52                                                & 0.57                                                & 0.36 \\                                         
\hline
\end{tabular}
    \caption{Cross-domain and cross-method experiments, using a hard-loss vs. a soft-loss. Columns show train setting and rows test performance. Acc. is for predicting the majority label. JDS compares predicted distribution (soft) with target distribution. * indicates cross-method results are not statistically significant (McNemar's test).
    }
    \label{tab:parser_experiment_single}
\end{table*}
\subsection{Case 2: cross-domain vs cross-method}
The purpose of this study is to investigate how cross-domain generalization is affected by method bias.
In other words, we want to compare a \textit{cross-domain and cross-method} setup with a \textit{cross-domain and same-method} setup. 
We test on the domain-specific data from the 1,200 instances annotated by QA and DC respectively and train on various domain configurations from DiscoGem (excluding dev and test), together with the extra 300 PDTB instances, annotated by DC. 

 Table \ref{tab:parser_experiment_single} shows the different combinations of data sets we use in this study (columns) as well as the results of in- and cross-domain and in- and cross-method predictions (rows). Both a change in domain and a change in annotation task lead to lower performance.
 Interestingly, the results show that the task factor has a stronger effect on performance than the domain: When training on DC distributions, the QA test results are worse than the DC test results in all cases. This indicates that task bias is an important factor to consider when training models.
Generally, except in the out-of-domain novel test case, training with a soft loss leads to the same or considerably better generalization accuracy than training with a hard loss. We thus confirm the findings of \citet{peterson2019human} and \citet{uma2020case} also for DR classification.

\section{Discussion and Conclusion}
DR annotation is a notoriously difficult task with low IAA. Annotations are not only subject to the interpretation of the coder \cite{spooren2010}, but also to the framework \cite{demberg2019compatible}.
The current study extends these findings by showing that the task design also crucially affects the output. 
We investigated the effect of two distinct crowdsourced DR annotation tasks on the obtained relation distributions. 
These two tasks are unique in that they use natural language to annotate. Even though these designs are more intuitive to laymen, we show that also such natural language-based annotation
designs suffer from bias and leave room for varying interpretations (as do traditional annotation tasks).

The results show that both methods have unique biases, but also that both methods are valid, as similar sets of labels are produced. Further, the methods seem to be complementary: both methods show higher agreement with the reference label than with each other. This indicates that the methods capture different sense types. The results further show that the textual domain can push each method towards different label distributions. Lastly we simulated how aggregating annotations based on method bias improves agreement.

We suggest several modifications to both methods for future work.
For QA, we recommend to replace question prefix options which start with a connective, such as "After what". The revised options should ideally start with a Wh-question word, for ex. "What happens after..". This would make the questions sound more natural and help to prevent confusion with respect to level-3 sense distinctions. For DC, an improved interface that allows workers to highlight argument spans could serve as a screen that confirms the relation is between the two consecutive sentences.
Syntactic constraints making it difficult to insert certain rare connectives could also be mitigated if the workers are allowed to make minor edits to the texts.

Considering that both methods show benefits and possible downsides, it could be interesting to combine them for future crowdsourcing efforts. Given that obtaining DC annotations is cheaper and quicker, it could make sense to collect DC annotations on a larger scale and then use the QA method for a specific subset that shows high label entropy. Another option would be to merge both methods, by first letting the crowdworkers insert a connective and then use QAs for the second connective-disambiguation step. Lastly, since we showed that often more than one relation sense can hold, it would make sense to allow annotators to write multiple QA pairs or insert multiple possible connectives for a given relation.

The DR classification experiments revealed that generalization across data from different task designs is hard, in the DC and QA case even harder than cross-domain generalization. Additionally, we found that merging data distributions coming from different task designs can help boost performance on data coming from a third source (traditional annotations). Lastly, we confirmed that soft modeling approaches using label distributions can improve discourse classification performance.

Task design bias has been identified as one source of annotation bias and acknowledged as an artifact of the dataset in other linguistic tasks as well \cite{pavlick2019inherent,jiang2022investigating}. Our findings show that the effect of this type of bias can be reduced by training with data collected by multiple methods. This could be the same for other NLP tasks, especially those cast in natural language, and comparing their task designs could be an interesting future research direction. We therefore encourage researchers to be more conscious about the biases crowdsourcing task design introduces.

\section*{Acknowledgements}
 This work was supported by the Deutsche Forschungsgemeinschaft, Funder Id: http://dx.doi.org/10.13039/501100001659, Grant Number: SFB1102: Information Density and Linguistic Encoding, by the the European Research Council, ERC-StG Grant no.\ 677352, and the Israel Science Foundation grant 2827/21, for which we are grateful. We also thank the TACL reviewers and Action Editors for their thoughtful comments.

\bibliography{tacl2021}

\begin{thebibliography}{73}
\expandafter\ifx\csname natexlab\endcsname\relax\def\natexlab#1{#1}\fi

\bibitem[{Aralikatte et~al.(2021)Aralikatte, Lamm, Hardt, and
  S{\o}gaard}]{aralikatte2019ellipsis}
Rahul Aralikatte, Matthew Lamm, Daniel Hardt, and Anders S{\o}gaard. 2021.
\newblock Ellipsis resolution as question answering: An evaluation.
\newblock In \emph{16th conference of the European Chapter of the Association
  for Computational Linguistics (EACL)}, pages 810--817. Association for
  Computational Linguistics.

\bibitem[{Aroyo and Welty(2013)}]{aroyo2013}
Lora Aroyo and Chris Welty. 2013.
\newblock {Crowd truth: Harnessing disagreement in crowdsourcing a relation
  extraction gold standard}.
\newblock \emph{WebSci2013 ACM}, 2013(2013).

\bibitem[{Artstein and Poesio(2008)}]{artstein2008}
Ron Artstein and Massimo Poesio. 2008.
\newblock Inter-coder agreement for computational linguistics.
\newblock \emph{Computational Linguistics}, 34(4):555--596.

\bibitem[{Asher(1993)}]{asher1993}
N.~Asher. 1993.
\newblock \emph{Reference to Abstract Objects in Discourse}.
\newblock Kluwer, Norwell, MA, Dordrecht.

\bibitem[{Basile et~al.(2021)Basile, Fell, Fornaciari, Hovy, Paun, Plank,
  Poesio, and Uma}]{Basile2021WeNT}
Valerio Basile, Michael Fell, Tommaso Fornaciari, Dirk Hovy, Silviu Paun,
  Barbara Plank, Massimo Poesio, and Alexandra Uma. 2021.
\newblock We need to consider disagreement in evaluation.
\newblock In \emph{BPPF}.

\bibitem[{Bowman et~al.(2015)Bowman, Angeli, Potts, and
  Manning}]{bowman2015large}
Samuel Bowman, Gabor Angeli, Christopher Potts, and Christopher~D Manning.
  2015.
\newblock A large annotated corpus for learning natural language inference.
\newblock In \emph{Proceedings of the 2015 Conference on Empirical Methods in
  Natural Language Processing}, pages 632--642.

\bibitem[{Bowman and Dahl(2021)}]{bowman2021will}
Samuel Bowman and George Dahl. 2021.
\newblock What will it take to fix benchmarking in natural language
  understanding?
\newblock In \emph{Proceedings of the 2021 Conference of the North American
  Chapter of the Association for Computational Linguistics: Human Language
  Technologies}, pages 4843--4855.

\bibitem[{Buechel and Hahn(2017{\natexlab{a}})}]{buechel2017emobank}
Sven Buechel and Udo Hahn. 2017{\natexlab{a}}.
\newblock Emobank: Studying the impact of annotation perspective and
  representation format on dimensional emotion analysis.
\newblock In \emph{Proceedings of the 15th Conference of the European Chapter
  of the Association for Computational Linguistics: Volume 2, Short Papers},
  pages 578--585.

\bibitem[{Buechel and Hahn(2017{\natexlab{b}})}]{buechel2017readers}
Sven Buechel and Udo Hahn. 2017{\natexlab{b}}.
\newblock Readers vs. writers vs. texts: Coping with different perspectives of
  text understanding in emotion annotation.
\newblock In \emph{Proceedings of the 11th Linguistic Annotation Workshop},
  pages 1--12.

\bibitem[{Carlson and Marcu(2001)}]{carlson2001}
Lynn Carlson and Daniel Marcu. 2001.
\newblock Discourse tagging reference manual.
\newblock \emph{ISI Technical Report ISI-TR-545}, 54:1--56.

\bibitem[{Chang et~al.(2016)Chang, Lee-Goldman, and
  Tseng}]{chang2016linguistic}
Nancy Chang, Russell Lee-Goldman, and Michael Tseng. 2016.
\newblock Linguistic wisdom from the crowd.
\newblock In \emph{Third AAAI Conference on Human Computation and
  Crowdsourcing}.

\bibitem[{Chen et~al.(2020)Chen, Jiang, Poliak, Sakaguchi, and
  Van~Durme}]{chen2019uncertain}
Tongfei Chen, Zheng~Ping Jiang, Adam Poliak, Keisuke Sakaguchi, and Benjamin
  Van~Durme. 2020.
\newblock Uncertain natural language inference.
\newblock In \emph{Proceedings of the 58th Annual Meeting of the Association
  for Computational Linguistics}, pages 8772--8779.

\bibitem[{Chung et~al.(2019)Chung, Song, Kutty, Hong, Kim, and
  Lasecki}]{chung2019efficient}
John Joon~Young Chung, Jean~Y Song, Sindhu Kutty, Sungsoo Hong, Juho Kim, and
  Walter~S Lasecki. 2019.
\newblock Efficient elicitation approaches to estimate collective crowd
  answers.
\newblock \emph{Proceedings of the ACM on Human-Computer Interaction},
  3(CSCW):1--25.

\bibitem[{Cohen(1960)}]{cohen1960}
Jacob Cohen. 1960.
\newblock A coefficient of agreement for nominal scales.
\newblock \emph{Educational and Psychological Measurement}, 20(1):37--46.

\bibitem[{Cowen et~al.(2019)Cowen, Sauter, Tracy, and
  Keltner}]{cowen2019mapping}
Alan Cowen, Disa Sauter, Jessica~L Tracy, and Dacher Keltner. 2019.
\newblock Mapping the passions: Toward a high-dimensional taxonomy of emotional
  experience and expression.
\newblock \emph{Psychological Science in the Public Interest}, 20(1):69--90.

\bibitem[{De~Marneffe et~al.(2012)De~Marneffe, Manning, and Potts}]{de2012did}
Marie-Catherine De~Marneffe, Christopher~D Manning, and Christopher Potts.
  2012.
\newblock Did it happen? the pragmatic complexity of veridicality assessment.
\newblock \emph{Computational linguistics}, 38(2):301--333.

\bibitem[{Demberg et~al.(2019)Demberg, Scholman, and
  Asr}]{demberg2019compatible}
Vera Demberg, Merel~CJ Scholman, and Fatemeh~Torabi Asr. 2019.
\newblock How compatible are our discourse annotation frameworks? insights from
  mapping rst-dt and pdtb annotations.
\newblock \emph{Dialogue \& Discourse}, 10(1):87--135.

\bibitem[{D{\'\i}az et~al.(2018)D{\'\i}az, Johnson, Lazar, Piper, and
  Gergle}]{diaz2018addressing}
Mark D{\'\i}az, Isaac Johnson, Amanda Lazar, Anne~Marie Piper, and Darren
  Gergle. 2018.
\newblock Addressing age-related bias in sentiment analysis.
\newblock In \emph{Proceedings of the 2018 chi conference on human factors in
  computing systems}, pages 1--14.

\bibitem[{Dumitrache(2015)}]{dumitrache2015crowdsourcing}
Anca Dumitrache. 2015.
\newblock Crowdsourcing disagreement for collecting semantic annotation.
\newblock In \emph{European Semantic Web Conference}, pages 701--710. Springer.

\bibitem[{Dumitrache et~al.(2018)Dumitrache, Inel, Aroyo, Timmermans, and
  Welty}]{dumitrache2018}
Anca Dumitrache, Oana Inel, Lora Aroyo, Benjamin Timmermans, and Chris Welty.
  2018.
\newblock {CrowdTruth 2.0: Quality metrics for crowdsourcing with
  disagreement}.
\newblock In \emph{1st Workshop on Subjectivity, Ambiguity and Disagreement in
  Crowdsourcing, and Short Paper 1st Workshop on Disentangling the Relation
  Between Crowdsourcing and Bias Management, SAD+ CrowdBias 2018}, pages
  11--18. CEUR-WS.

\bibitem[{Dumitrache et~al.(2021)Dumitrache, Inel, Timmermans, Ortiz, Sips,
  Aroyo, and Welty}]{dumitrache2021empirical}
Anca Dumitrache, Oana Inel, Benjamin Timmermans, Carlos Ortiz, Robert-Jan Sips,
  Lora Aroyo, and Chris Welty. 2021.
\newblock Empirical methodology for crowdsourcing ground truth.
\newblock \emph{Semantic Web}, 12(3):403--421.

\bibitem[{Elazar et~al.(2022)Elazar, Basmov, Goldberg, and
  Tsarfaty}]{elazar2022text}
Yanai Elazar, Victoria Basmov, Yoav Goldberg, and Reut Tsarfaty. 2022.
\newblock Text-based np enrichment.
\newblock \emph{Transactions of the Association for Computational Linguistics},
  10:764--784.

\bibitem[{Erk and McCarthy(2009)}]{erk2009graded}
Katrin Erk and Diana McCarthy. 2009.
\newblock Graded word sense assignment.
\newblock In \emph{Proceedings of the 2009 conference on empirical methods in
  natural language processing}, pages 440--449.

\bibitem[{Ferracane et~al.(2021)Ferracane, Durrett, Li, and
  Erk}]{ferracane2021did}
Elisa Ferracane, Greg Durrett, Junyi~Jessy Li, and Katrin Erk. 2021.
\newblock Did they answer? subjective acts and intents in conversational
  discourse.
\newblock In \emph{Proceedings of the 2021 Conference of the North American
  Chapter of the Association for Computational Linguistics: Human Language
  Technologies}, pages 1626--1644.

\bibitem[{Fitzgerald et~al.(2018)Fitzgerald, Michael, He, and
  Zettlemoyer}]{fitzgerald2018large}
Nicholas Fitzgerald, Julian Michael, Luheng He, and Luke Zettlemoyer. 2018.
\newblock Large-scale qa-srl parsing.
\newblock In \emph{Proceedings of the 56th Annual Meeting of the Association
  for Computational Linguistics (Volume 1: Long Papers)}, pages 2051--2060.

\bibitem[{He et~al.(2020)He, Liu, Gao, and Chen}]{he2020deberta}
Pengcheng He, Xiaodong Liu, Jianfeng Gao, and Weizhu Chen. 2020.
\newblock Deberta: Decoding-enhanced bert with disentangled attention.
\newblock In \emph{International Conference on Learning Representations}.

\bibitem[{Hou(2020)}]{hou2020bridging}
Yufang Hou. 2020.
\newblock Bridging anaphora resolution as question answering.
\newblock In \emph{Proceedings of the 58th Annual Meeting of the Association
  for Computational Linguistics}, pages 1428--1438.

\bibitem[{Hovy et~al.(2013)Hovy, Berg-Kirkpatrick, Vaswani, and
  Hovy}]{hovy2013}
Dirk Hovy, Taylor Berg-Kirkpatrick, Ashish Vaswani, and Eduard Hovy. 2013.
\newblock {Learning whom to trust with MACE}.
\newblock In \emph{Proceedings of the 2013 Conference of the North American
  Chapter of the Association for Computational Linguistics: Human Language
  Technologies (NAACL-HLT)}, pages 1120--1130, Denver, CO.

\bibitem[{Hube et~al.(2019)Hube, Fetahu, and Gadiraju}]{hube2019understanding}
Christoph Hube, Besnik Fetahu, and Ujwal Gadiraju. 2019.
\newblock Understanding and mitigating worker biases in the crowdsourced
  collection of subjective judgments.
\newblock In \emph{Proceedings of the 2019 CHI Conference on Human Factors in
  Computing Systems}, pages 1--12.

\bibitem[{Jakobsen et~al.(2022)Jakobsen, Barrett, S{\o}gaard, and
  Lassen}]{jakobsen2022sensitivity}
Terne Sasha~Thorn Jakobsen, Maria Barrett, Anders S{\o}gaard, and David Lassen.
  2022.
\newblock The sensitivity of annotator bias to task definitions in argument
  mining.
\newblock In \emph{Proceedings of the 16th Lingusitic Annotation Workshop
  (LAW-XVI) within LREC2022}, pages 44--61.

\bibitem[{Jiang and de~Marneffe(2022)}]{jiang2022investigating}
Nanjiang Jiang and Marie-Catherine de~Marneffe. 2022.
\newblock Investigating reasons for disagreement in natural language inference.
\newblock \emph{Transactions of the Association for Computational Linguistics},
  10:1357--1374.

\bibitem[{Jiang et~al.(2017)Jiang, Kummerfeld, and
  Lasecki}]{jiang2017understanding}
Youxuan Jiang, Jonathan~K Kummerfeld, and Walter Lasecki. 2017.
\newblock Understanding task design trade-offs in crowdsourced paraphrase
  collection.
\newblock In \emph{Proceedings of the 55th Annual Meeting of the Association
  for Computational Linguistics (Volume 2: Short Papers)}, pages 103--109.

\bibitem[{Jurgens(2013)}]{jurgens2013embracing}
David Jurgens. 2013.
\newblock Embracing ambiguity: A comparison of annotation methodologies for
  crowdsourcing word sense labels.
\newblock In \emph{Proceedings of the 2013 Conference of the North American
  Chapter of the Association for Computational Linguistics: Human Language
  Technologies}, pages 556--562.

\bibitem[{Kawahara et~al.(2014)Kawahara, Machida, Shibata, Kurohashi,
  Kobayashi, and Sassano}]{kawahara2014}
Daisuke Kawahara, Yuichiro Machida, Tomohide Shibata, Sadao Kurohashi, Hayato
  Kobayashi, and Manabu Sassano. 2014.
\newblock Rapid development of a corpus with discourse annotations using
  two-stage crowdsourcing.
\newblock In \emph{Proceedings of the International Conference on Computational
  Linguistics (COLING)}, pages 269--278, Dublin, Ireland.

\bibitem[{Kishimoto et~al.(2018)Kishimoto, Sawada, Murawaki, Kawahara, and
  Kurohashi}]{Kishimoto2018ImprovingCA}
Yudai Kishimoto, Shinnosuke Sawada, Yugo Murawaki, Daisuke Kawahara, and Sadao
  Kurohashi. 2018.
\newblock Improving crowdsourcing-based annotation of japanese discourse
  relations.
\newblock In \emph{LREC}.

\bibitem[{Ko et~al.(2021)Ko, Dalton, Simmons, Fisher, Durrett, and
  Li}]{ko2021discourse}
Wei-Jen Ko, Cutter Dalton, Mark Simmons, Eliza Fisher, Greg Durrett, and
  Junyi~Jessy Li. 2021.
\newblock Discourse comprehension: A question answering framework to represent
  sentence connections.
\newblock \emph{arXiv preprint arXiv:2111.00701}.

\bibitem[{Koehn(2005)}]{Koehn-Europarl-2005}
Philipp Koehn. 2005.
\newblock Europarl: A parallel corpus for statistical machine translation.
\newblock In \emph{Proceedings of MT Summit X}, pages 79--86, Phuket, Thailand.

\bibitem[{Luo et~al.(2020)Luo, Card, and Jurafsky}]{luo2020detecting}
Yiwei Luo, Dallas Card, and Dan Jurafsky. 2020.
\newblock Detecting stance in media on global warming.
\newblock In \emph{Findings of the Association for Computational Linguistics:
  EMNLP 2020}, pages 3296--3315.

\bibitem[{Mann and Thompson(1988)}]{mann1988}
William~C Mann and Sandra~A Thompson. 1988.
\newblock Rhetorical {Structure} {Theory}: Toward a functional theory of text
  organization.
\newblock \emph{Text-Interdisciplinary Journal for the Study of Discourse},
  8(3):243--281.

\bibitem[{Manning(2006)}]{manning20061}
Christopher~D Manning. 2006.
\newblock Local textual inference: It's hard to circumscribe, but you know it
  when you see it -- and nlp needs it.

\bibitem[{Marchal et~al.(2022)Marchal, Scholman, Yung, and
  Demberg}]{marchal2022establishing}
Marian Marchal, Merel Scholman, Frances Yung, and Vera Demberg. 2022.
\newblock Establishing annotation quality in multi-label annotations.
\newblock In \emph{Proceedings of the 29th International Conference on
  Computational Linguistics}, pages 3659--3668.

\bibitem[{Min et~al.(2020)Min, Michael, Hajishirzi, and
  Zettlemoyer}]{min2020ambigqa}
Sewon Min, Julian Michael, Hannaneh Hajishirzi, and Luke Zettlemoyer. 2020.
\newblock Ambigqa: Answering ambiguous open-domain questions.
\newblock In \emph{Proceedings of the 2020 Conference on Empirical Methods in
  Natural Language Processing (EMNLP)}, pages 5783--5797.

\bibitem[{Nie et~al.(2020)Nie, Zhou, and Bansal}]{nie2020can}
Yixin Nie, Xiang Zhou, and Mohit Bansal. 2020.
\newblock What can we learn from collective human opinions on natural language
  inference data?
\newblock In \emph{Proceedings of the 2020 Conference on Empirical Methods in
  Natural Language Processing (EMNLP)}, pages 9131--9143.

\bibitem[{Passonneau and Carpenter(2014)}]{passonneau2014}
Rebecca~J Passonneau and Bob Carpenter. 2014.
\newblock The benefits of a model of annotation.
\newblock \emph{Transactions of the Association for Computational Linguistics},
  2:311--326.

\bibitem[{Pavlick and Kwiatkowski(2019)}]{pavlick2019inherent}
Ellie Pavlick and Tom Kwiatkowski. 2019.
\newblock Inherent disagreements in human textual inferences.
\newblock \emph{Transactions of the Association for Computational Linguistics},
  7:677--694.

\bibitem[{Peterson et~al.(2019)Peterson, Battleday, Griffiths, and
  Russakovsky}]{peterson2019human}
Joshua~C Peterson, Ruairidh~M Battleday, Thomas~L Griffiths, and Olga
  Russakovsky. 2019.
\newblock Human uncertainty makes classification more robust.
\newblock In \emph{Proceedings of the IEEE/CVF International Conference on
  Computer Vision}, pages 9617--9626.

\bibitem[{Plank et~al.(2014)Plank, Hovy, and
  S{\o}gaard}]{plank2014linguistically}
Barbara Plank, Dirk Hovy, and Anders S{\o}gaard. 2014.
\newblock Linguistically debatable or just plain wrong?
\newblock In \emph{Proceedings of the 52nd Annual Meeting of the Association
  for Computational Linguistics (Volume 2: Short Papers)}, pages 507--511.

\bibitem[{Poesio and Artstein(2005)}]{poesio2005reliability}
Massimo Poesio and Ron Artstein. 2005.
\newblock The reliability of anaphoric annotation, reconsidered: Taking
  ambiguity into account.
\newblock In \emph{Proceedings of the workshop on frontiers in corpus
  annotations ii: Pie in the sky}, pages 76--83.

\bibitem[{Poesio et~al.(2006)Poesio, Sturt, Artstein, and
  Filik}]{poesio2006underspecification}
Massimo Poesio, Patrick Sturt, Ron Artstein, and Ruth Filik. 2006.
\newblock Underspecification and anaphora: Theoretical issues and preliminary
  evidence.
\newblock \emph{Discourse processes}, 42(2):157--175.

\bibitem[{Prabhakaran et~al.(2021)Prabhakaran, Davani, and
  Diaz}]{prabhakaran2021releasing}
Vinodkumar Prabhakaran, Aida~Mostafazadeh Davani, and Mark Diaz. 2021.
\newblock On releasing annotator-level labels and information in datasets.
\newblock In \emph{Proceedings of The Joint 15th Linguistic Annotation Workshop
  (LAW) and 3rd Designing Meaning Representations (DMR) Workshop}, pages
  133--138.

\bibitem[{Pyatkin et~al.(2020)Pyatkin, Klein, Tsarfaty, and
  Dagan}]{pyatkin2020}
Valentina Pyatkin, Ayal Klein, Reut Tsarfaty, and Ido Dagan. 2020.
\newblock {QADiscourse-Discourse Relations as QA Pairs}: Representation,
  crowdsourcing and baselines.
\newblock In \emph{Proceedings of the 2020 Conference on Empirical Methods in
  Natural Language Processing (EMNLP)}, pages 2804--2819.

\bibitem[{Rajpurkar et~al.(2018)Rajpurkar, Jia, and Liang}]{rajpurkar2018know}
Pranav Rajpurkar, Robin Jia, and Percy Liang. 2018.
\newblock Know what you don’t know: Unanswerable questions for squad.
\newblock In \emph{Proceedings of the 56th Annual Meeting of the Association
  for Computational Linguistics (Volume 2: Short Papers)}, pages 784--789.

\bibitem[{Rehbein et~al.(2016)Rehbein, Scholman, and Demberg}]{rehbein2016}
Ines Rehbein, Merel Scholman, and Vera Demberg. 2016.
\newblock \href {https://aclanthology.org/L16-1165} {Annotating discourse
  relations in spoken language: A comparison of the {PDTB} and {CCR}
  frameworks}.
\newblock In \emph{Proceedings of the Tenth International Conference on
  Language Resources and Evaluation ({LREC}'16)}, pages 1039--1046,
  Portoro{\v{z}}, Slovenia. European Language Resources Association (ELRA).

\bibitem[{Riezler(2014)}]{riezler2014}
Stefan Riezler. 2014.
\newblock On the problem of theoretical terms in empirical computational
  linguistics.
\newblock \emph{Computational Linguistics}, 40(1):235--245.

\bibitem[{Rohde et~al.(2016)Rohde, Dickinson, Schneider, Clark, Louis, and
  Webber}]{rohde2016}
Hannah Rohde, Anna Dickinson, Nathan Schneider, Christopher Clark, Annie Louis,
  and Bonnie Webber. 2016.
\newblock Filling in the blanks in understanding discourse adverbials:
  Consistency, conflict, and context-dependence in a crowdsourced elicitation
  task.
\newblock In \emph{Proceedings of the 10th Linguistic Annotation Workshop (LAW
  X)}, pages 49--58, Berlin, Germany.

\bibitem[{Sanders et~al.(1992)Sanders, Spooren, and Noordman}]{sanders1992}
Ted J.~M. Sanders, Wilbert P. M.~S. Spooren, and Leo G.~M. Noordman. 1992.
\newblock Toward a taxonomy of coherence relations.
\newblock \emph{Discourse Processes}, 15(1):1--35.

\bibitem[{Scholman and Demberg(2017)}]{scholman2017context}
Merel C~J Scholman and Vera Demberg. 2017.
\newblock Crowdsourcing discourse interpretations: On the influence of context
  and the reliability of a connective insertion task.
\newblock In \emph{Proceedings of the 11th Linguistic Annotation Workshop
  (LAW)}, pages 24--33, Valencia, Spain.

\bibitem[{Scholman et~al.(2022{\natexlab{a}})Scholman, Dong, Yung, and
  Demberg}]{scholman2022DiscoGeM}
Merel C.~J. Scholman, Tianai Dong, Frances Yung, and Vera Demberg.
  2022{\natexlab{a}}.
\newblock Discogem: A crowdsourced corpus of genre-mixed implicit discourse
  relations.
\newblock In \emph{Proceedings of the Thirteenth International Conference on
  Language Resources and Evaluation ({LREC}'22)}, Marseille, France. European
  Language Resources Association (ELRA).

\bibitem[{Scholman et~al.(2022{\natexlab{b}})Scholman, Pyatkin, Yung, Dagan,
  Tsarfaty, and Demberg}]{scholman2022design}
Merel C.~J. Scholman, Valentina Pyatkin, Frances Yung, Ido Dagan, Reut
  Tsarfaty, and Vera Demberg. 2022{\natexlab{b}}.
\newblock Design choices in crowdsourcing discourse relation annotations: The
  effect of worker selection and training.
\newblock In \emph{Proceedings of the Thirteenth International Conference on
  Language Resources and Evaluation ({LREC}'22)}, Marseille, France. European
  Language Resources Association (ELRA).

\bibitem[{Shi and Demberg(2019)}]{shi2019}
Wei Shi and Vera Demberg. 2019.
\newblock \href {https://doi.org/10.18653/v1/W19-0416} {Learning to explicitate
  connectives with {S}eq2{S}eq network for implicit discourse relation
  classification}.
\newblock In \emph{Proceedings of the 13th International Conference on
  Computational Semantics - Long Papers}, pages 188--199, Gothenburg, Sweden.
  Association for Computational Linguistics.

\bibitem[{Snow et~al.(2008)Snow, O'Connor, Jurafsky, and Ng}]{snow2008}
Rion Snow, Brendan O'Connor, Daniel Jurafsky, and Andrew~Y Ng. 2008.
\newblock Cheap and fast---but is it good? evaluating non-expert annotations
  for natural language tasks.
\newblock In \emph{Proceedings of the Conference on Empirical Methods in
  Natural Language Processing (EMNLP)}, pages 254--263, Waikiki, HI.

\bibitem[{Spooren and Degand(2010)}]{spooren2010}
Wilbert P. M.~S. Spooren and Liesbeth Degand. 2010.
\newblock Coding coherence relations: Reliability and validity.
\newblock \emph{Corpus Linguistics and Linguistic Theory}, 6(2):241--266.

\bibitem[{Uma et~al.(2020)Uma, Fornaciari, Hovy, Paun, Plank, and
  Poesio}]{uma2020case}
Alexandra Uma, Tommaso Fornaciari, Dirk Hovy, Silviu Paun, Barbara Plank, and
  Massimo Poesio. 2020.
\newblock A case for soft loss functions.
\newblock In \emph{Proceedings of the AAAI Conference on Human Computation and
  Crowdsourcing}, volume~8, pages 173--177.

\bibitem[{Uma et~al.(2021)Uma, Fornaciari, Hovy, Paun, Plank, and
  Poesio}]{uma2021learning}
Alexandra~N Uma, Tommaso Fornaciari, Dirk Hovy, Silviu Paun, Barbara Plank, and
  Massimo Poesio. 2021.
\newblock Learning from disagreement: A survey.
\newblock \emph{Journal of Artificial Intelligence Research}, 72:1385--1470.

\bibitem[{Waseem(2016)}]{waseem2016you}
Zeerak Waseem. 2016.
\newblock Are you a racist or am i seeing things? annotator influence on hate
  speech detection on twitter.
\newblock In \emph{Proceedings of the first workshop on NLP and computational
  social science}, pages 138--142.

\bibitem[{Webber(2009)}]{webber2009}
Bonnie Webber. 2009.
\newblock {Genre distinctions for discourse in the Penn TreeBank}.
\newblock In \emph{Proceedings of the Joint Conference of the 47th Annual
  Meeting of the ACL and the 4th International Joint Conference on Natural
  Language Processing of the AFNLP}, pages 674--682.

\bibitem[{Webber et~al.(2019)Webber, Prasad, Lee, and Joshi}]{webber2019}
Bonnie Webber, Rashmi Prasad, Alan Lee, and Aravind Joshi. 2019.
\newblock {The Penn Discourse Treebank 3.0 annotation manual}.
\newblock \emph{Philadelphia, University of Pennsylvania}.

\bibitem[{Wolf et~al.(2020)Wolf, Debut, Sanh, Chaumond, Delangue, Moi, Cistac,
  Rault, Louf, Funtowicz et~al.}]{wolf2019huggingface}
Thomas Wolf, Lysandre Debut, Victor Sanh, Julien Chaumond, Clement Delangue,
  Anthony Moi, Pierric Cistac, Tim Rault, R{\'e}mi Louf, Morgan Funtowicz,
  et~al. 2020.
\newblock Transformers: State-of-the-art natural language processing.
\newblock In \emph{Proceedings of the 2020 conference on empirical methods in
  natural language processing: system demonstrations}, pages 38--45.

\bibitem[{Yung et~al.(2019)Yung, Demberg, and Scholman}]{yung2019}
Frances Yung, Vera Demberg, and Merel Scholman. 2019.
\newblock Crowdsourcing discourse relation annotations by a two-step connective
  insertion task.
\newblock In \emph{Proceedings of the 13th Linguistic Annotation Workshop},
  pages 16--25.

\bibitem[{Zeyrek et~al.(2019)Zeyrek, Mendes, Grishina, Kurfal{\i}, Gibbon, and
  Ogrodniczuk}]{zeyrek2019}
Deniz Zeyrek, Am{\'a}lia Mendes, Yulia Grishina, Murathan Kurfal{\i}, Samuel
  Gibbon, and Maciej Ogrodniczuk. 2019.
\newblock Ted multilingual discourse bank (ted-mdb): a parallel corpus
  annotated in the pdtb style.
\newblock \emph{Language Resources and Evaluation}, pages 1--27.

\bibitem[{Zeyrek et~al.(2020)Zeyrek, Mendes, Grishina, Kurfal{\i}, Gibbon, and
  Ogrodniczuk}]{zeyrek2020ted}
Deniz Zeyrek, Am{\'a}lia Mendes, Yulia Grishina, Murathan Kurfal{\i}, Samuel
  Gibbon, and Maciej Ogrodniczuk. 2020.
\newblock Ted multilingual discourse bank (ted-mdb): a parallel corpus
  annotated in the pdtb style.
\newblock \emph{Language Resources and Evaluation}, 54(2):587--613.

\bibitem[{Zhang et~al.(2021)Zhang, Gong, and Choi}]{zhang2021learning}
Shujian Zhang, Chengyue Gong, and Eunsol Choi. 2021.
\newblock Learning with different amounts of annotation: From zero to many
  labels.
\newblock In \emph{Proceedings of the 2021 Conference on Empirical Methods in
  Natural Language Processing}, pages 7620--7632.

\bibitem[{Zik{\'a}nov{\'a} et~al.(2019)Zik{\'a}nov{\'a}, M{\'\i}rovsk{\`y}, and
  Synkov{\'a}}]{zikanova2019explicit}
{\v{S}}{\'a}rka Zik{\'a}nov{\'a}, Ji{\v{r}}{\'\i} M{\'\i}rovsk{\`y}, and
  Pavl{\'\i}na Synkov{\'a}. 2019.
\newblock Explicit and implicit discourse relations in the {Prague Discourse
  Treebank}.
\newblock In \emph{Text, Speech, and Dialogue: 22nd International Conference,
  TSD 2019, Ljubljana, Slovenia, September 11--13, 2019, Proceedings 22}, pages
  236--248. Springer.

\end{thebibliography}
\bibliographystyle{acl_natbib}

\end{document}